\relax
\documentclass[letterpaper]{article} 
\usepackage{aaai20}  
\usepackage{times}  
\usepackage{helvet} 
\usepackage{courier}  
\usepackage[hyphens]{url}  
\usepackage{graphicx} 
\usepackage{graphicx}  
\usepackage{subcaption}
\title{A Study on Multimodal and Interactive Explanations for Visual Question Answering}
\author{Kamran Alipour,\textsuperscript{1}\thanks{Email: kalipour@eng.ucsd.edu} Jurgen P. Schulze,\textsuperscript{1} Yi Yao,\textsuperscript{2} Avi Ziskind,\textsuperscript{2} and Giedrius Burachas\textsuperscript{2} \\
\textsuperscript{1}{UC San Diego, La Jolla, CA} \\
\textsuperscript{2}{SRI International, Princeton, NJ}
}
\begin{document}

\maketitle
\begin{abstract}
Explainability and interpretability of AI models is an essential factor affecting the safety of AI. While various explainable AI (XAI) approaches aim at mitigating the lack of transparency in deep networks, the evidence of the effectiveness of these approaches in improving usability, trust, and understanding of AI systems are still missing. We evaluate multimodal explanations in the setting of a Visual Question Answering (VQA) task, by asking users to predict the response accuracy of a VQA agent with and without explanations. We use between-subjects and within-subjects experiments to probe explanation effectiveness in terms of improving user prediction accuracy, confidence, and reliance, among other factors. The results indicate that the explanations help improve human prediction accuracy, especially in trials when the VQA system's answer is inaccurate. Furthermore, we introduce active attention, a novel method for evaluating causal attentional effects through intervention by editing attention maps. User explanation ratings are strongly correlated with human prediction accuracy and suggest the efficacy of these explanations in human-machine AI collaboration tasks.
\end{abstract}

\section{Introduction}

With recent developments in deep learning models and access to ever increasing data in all fields, we have witnessed a growing interest in using neural networks in a variety of applications over the past several years. Many complex tasks which required manual human effort are now assigned to these AI systems.
To utilize an AI system effectively, users need a basic understanding of the system, i.e., they need to build a mental model of the system's operation for anticipating success and failure modes, and to develop a certain level of trust in that system. However, deep learning models are notoriously opaque and difficult to interpret and often have unexpected failure modes, making it hard to build trust. AI systems which users do not understand and trust are impractical for most applications, especially where vital decisions are made based on AI results. Previous efforts to address this issue and explain the inner workings of deep learning models include visualizing intermediate features of importance \cite{zeiler2014visualizing,zhou2014object,selvaraju2017grad} and providing textual justifications \cite{huk2018multimodal}, but these studies did not evaluate whether these explanations aided human users in better understanding the system inferences or if they helped build trust. Prior work has quantified the effectiveness of their explanations by collecting user ratings \cite{lu2016hierarchical,chandrasekaran2017takes} or checking their alignment with human attention \cite{das2017human}, but found no substantial benefit for the explanation types used in that study.

To promote understanding of and trust in the system, we propose an approach that provides transparency about the intermediate stages of the model operation, such as attentional masks and detected/attended objects in the scene. Also, we generate textual explanations that are aimed to explain \emph{why} a particular answer was generated. Our explanations fall under the category of \textit{local explanations} as they are intended to address inference on a specific run of the VQA system and are valid for that run. We offer extensive evaluations of these explanations in the setting of a VQA system. These evaluations are made by human subjects while performing a correctness prediction task.  After seeing an image, a question, and some explanations, subjects are asked to predict whether the explainable VQA (XVQA) system will be accurate or not. We collect both the data on subject prediction performance and their explanation ratings during and after each prediction run.   

We also introduce active attention - an interactive approach to explaining answers from a VQA system. We provide an interactive framework to deploy this new explanation mode. The interface is used to conduct a user study on the \textit{effectiveness} and \textit{helpfulness} of explanations in terms of improving user's performance in user-machine tasks and also their mental model of the system. The efficacy of explanations is measured using several metrics described below. We show that explanations improve VQA correctness prediction performance on runs with incorrect answers, thus indicating that explanations are very effective in anticipating VQA failure. Explanations rated as more helpful are more likely to help predict VQA outcome correctly. Interestingly, the user confidence in their prediction exhibits substantial correlation with the VQA system confidence (top answer probability). This finding further supports the notion that the subjects develop a mental model of the XQA system that helps them judge when to trust the system and when not.

\section{Related Work}

\textbf{Visual Question Answering.} In the VQA task, the system provides a question and an image, and the task is to answer the question using the image correctly. The multimodal aspect of the problem, combining both natural language and visual features makes this a challenging task. The VQA problem was originally introduced in \cite{antol2015vqa} and since then, multiple variations have been proposed and tested. A common approach is to use attentional masks that highlight specific regions of the image, conditioned on the question \cite{DBLP:journals/corr/KazemiE17,DBLP:journals/corr/LuYBP16,DBLP:journals/corr/abs-1708-02711,DBLP:journals/corr/XuS15a,DBLP:journals/corr/abs-1807-09956,fukui2016multimodal,10.1007/978-3-319-46478-7_28,teney2018tips}.

\textbf{Explainable AI.}  The effort to produce automated reasoning and explanations dates back to very early work in the AI field with direct applications in medicine \cite{shortliffe1984model}, education \cite{lane2005explainable,van2004explainable}, and robotics \cite{lomas2012explaining}.
For vision-based AI applications, several explanation systems draw the focus on discovering visual features important in the decision-making process \cite{zeiler2014visualizing,hendricks2016generating,jiang2017learning,selvaraju2017grad,jiang2018pythia}. For visual question answering tasks, explanations usually involve image or language attention \cite{lu2016hierarchical,DBLP:journals/corr/KazemiE17}. Besides saliency/attention maps, other work has studied different explanation modes including layered attentions \cite{yang2016stacked}, bounding boxes around important regions \cite{anne2018grounding} or textual justifications \cite{shortliffe1984model,huk2018multimodal}.\\  In this paper, we propose a multi-modal explanation system which includes justifications for system behavior in visual, textual, and semantic formats. Unlike previous work that suggest explanations mostly relied on information produced by the AI machine, our approach benefits from combining AI-generated explanations and human-annotations for better interpretability.



\textbf{Human studies.}
As an attempt to assess the role of an explanation system in building a better mental model of AI systems for their human users, several previous efforts focused on quantifying the efficacy of explanations through user studies. Some of these studies were developed around measuring trust with users \cite{cosley2003seeing,ribeiro2016should}, or the role of explanations to achieve a goal \cite{kulesza2012tell,narayanan2018humans,ray2019lucid}. Other works measured the quality of explanations based on improving the predictability of a VQA model \cite{chandrasekaran2018explanations}.\\
Despite their great insights into the efficacy of various explanation modes, previous studies do not interactively involve the human subjects in producing these explanations. In our study, we design an interactive environment for users to evaluate our multi-modal explanation system in helping users predict the correctness of a VQA model. Moreover, The users also take part in generating explanations and receive online feedback from the AI machine.

\section{The VQA Model}
VQA deep learning models are trained to take an image and a question about its content and produce the answer to the question. The core model extracts features from natural language questions as well as images, combines them, and generates a natural language answer. Among various methods to train VQA systems to accomplish this task, the attention-based approach is specifically of our interest. 

\begin{figure}[ht]
    \centering
    \includegraphics[width=1.0\columnwidth]{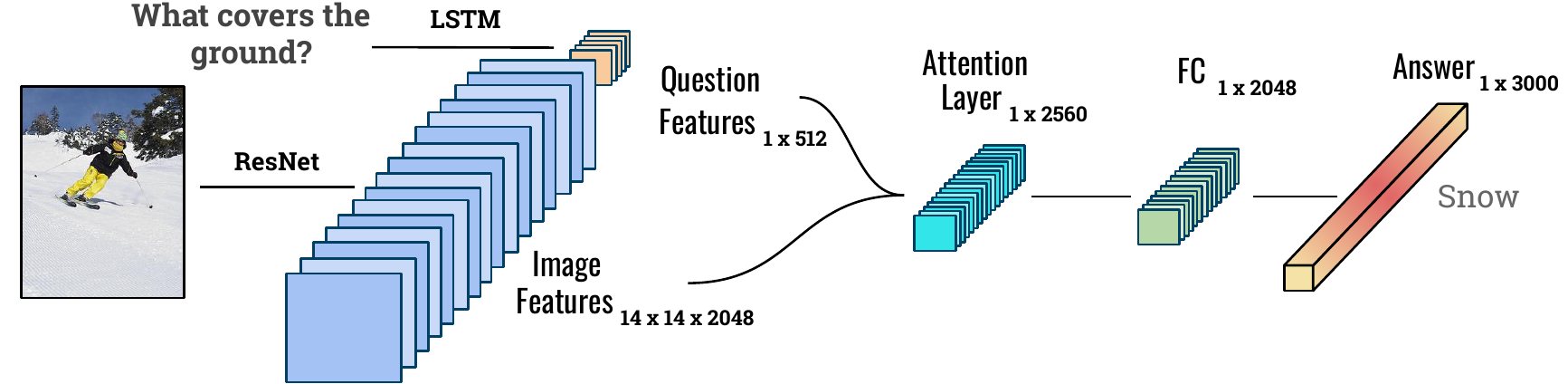}
    \caption{2017 SOTA VQA Architecture.}
    \label{fig:SOTA}
\end{figure}

We use a 2017 SOTA VQA model with a ResNet \cite{szegedy2017inception} image encoder (figure \ref{fig:SOTA}) as our VQA agent. The model is trained on VQA2 dataset and uses an attention mechanism to select visual features generated by an image encoder and an answer classifier that predicts an answer from 3000 candidates. Moreover, we replaced Resnet with a Mask-RCNN \cite{He_2017_ICCV} encoder to produce object attention explanations (similar to the approach used by \cite{ray2019lucid}).

As illustrated in figure \ref{fig:SOTA}, our VQA model takes as input a $224 \times 224$ RGB image and question with at most 15 words. Using a ResNet, the model encodes the image to reach a $14 \times 14 \times 2048$ feature representation. The model encodes the input question to a feature vector of 512 dimensions using an LSTM model based on the GloVe \cite{pennington2014glove} embedding of the words. The attention layer takes in the question and image feature representations and outputs a set of weights to attend on the image features. The weighted image features, concatenated with the question representation, is used to predict the final answer from a set of 3000 answer choices.

\section{Explanation Modes}

Our XVQA system aims at explaining the VQA agent's behavior by combining the attention features generated in the VQA model with meaningful annotations from the input data. These annotations include labels, descriptions, and bounding boxes of entities in the scene and their connections with each other.

Our XVQA model either visualizes information from the inner layers of the VQA model or incorporates that information with annotations to explain model's inner work. The explanations are provided in different combinations to the subgroups of study participants to assess their effectiveness for accurate prediction.
\subsection{Spatial attention}
\begin{figure}[ht]
    \centering
    \includegraphics[width=1.0\columnwidth]{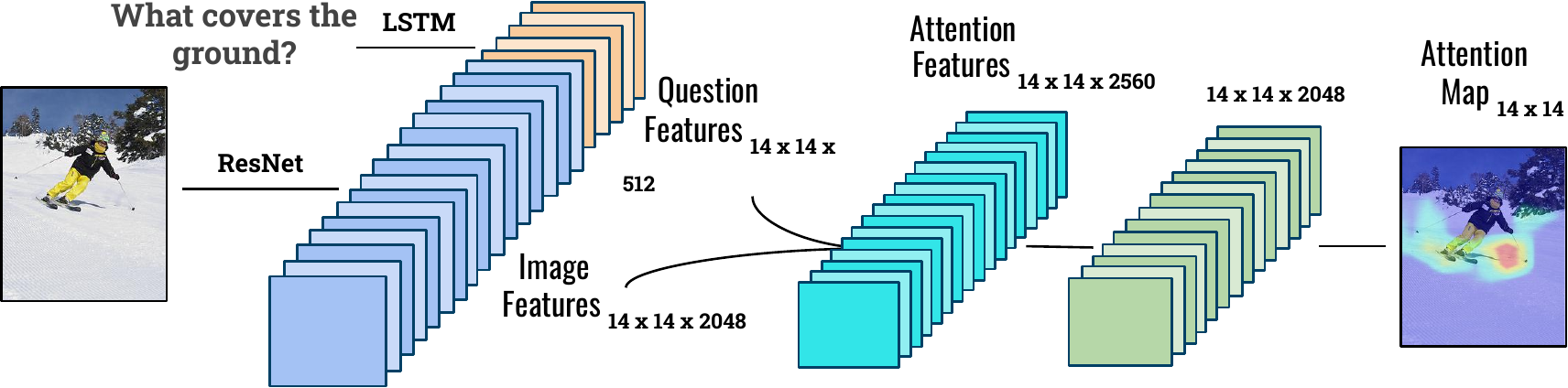}
    \caption{Attention map generated based on the input features in XVQA model.}
    \label{fig:VQA_Attention_Explanation_Model_Architecture}
\end{figure}

As introduced by previous work, the primary purpose of spatial attention is to show the parts of the image the model focuses on while preparing the answer. Attentions maps here are question-guided and more weighted in the areas of the image that make a higher contribution in the response generated by the model. The model computes the attentions based on image features in ResNet \cite{szegedy2017inception} layers and the question input. The final values in the attention map is a nonlinear function of image and question feature channels (figure \ref{fig:VQA_Attention_Explanation_Model_Architecture}) 

\begin{figure}[ht]
\centering
\begin{tabular}{@{}c@{}}
  \centering
  \includegraphics[width=.45\columnwidth]{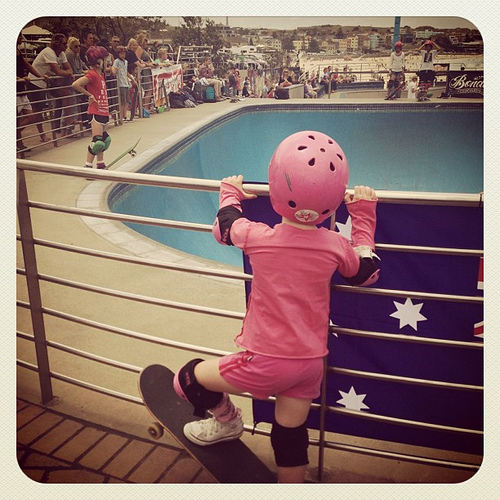}
  \label{fig:spatialinput}
 \end{tabular}
\begin{tabular}{@{}c@{}}
  \centering
  \includegraphics[width=.45\columnwidth]{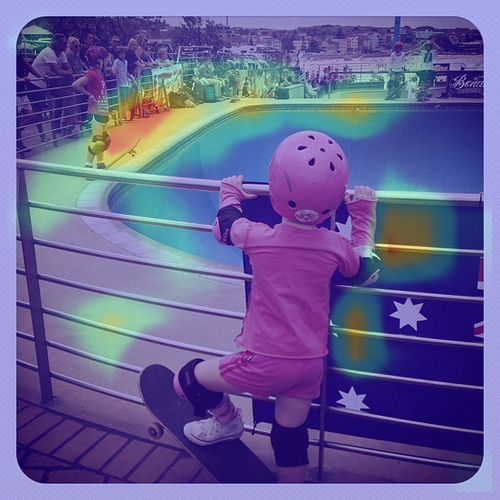}
 \end{tabular}
    \caption{Spatial attention explanation generated for the question: "What are the girls skating inside of?".}
     \label{fig:spatialoutput}
\end{figure}

Users try to develop an understanding of the way the model analyzes an image based on the question by looking at these attentions maps (example provided in figure \ref{fig:spatialoutput}).

\subsection{Active attention}
Our model provides this explanation mode for the users within a feedback loop. Users can utilize this feature to \textit{alter} a model's attention map to \textit{steer} the model's attention and the way the answer is generated. In this feedback loop, users first see the model's answer based on the original attention map, and then they modify the attention to create a different response.

The active attention trial has a two-step task to complete. The first step is very similar to spatial attention trials where users make their prediction based on the attention map generated by the VQA model. The subject then observes the prediction results and realizes whether the system is accurate or not. At the second step, the subject is asked to draw a new attention map. Using the manually drawn attention map, the model processes the image and question one more time and produces a second answer. 

\begin{figure}[ht]
\centering
  \includegraphics[width=1.0\columnwidth]{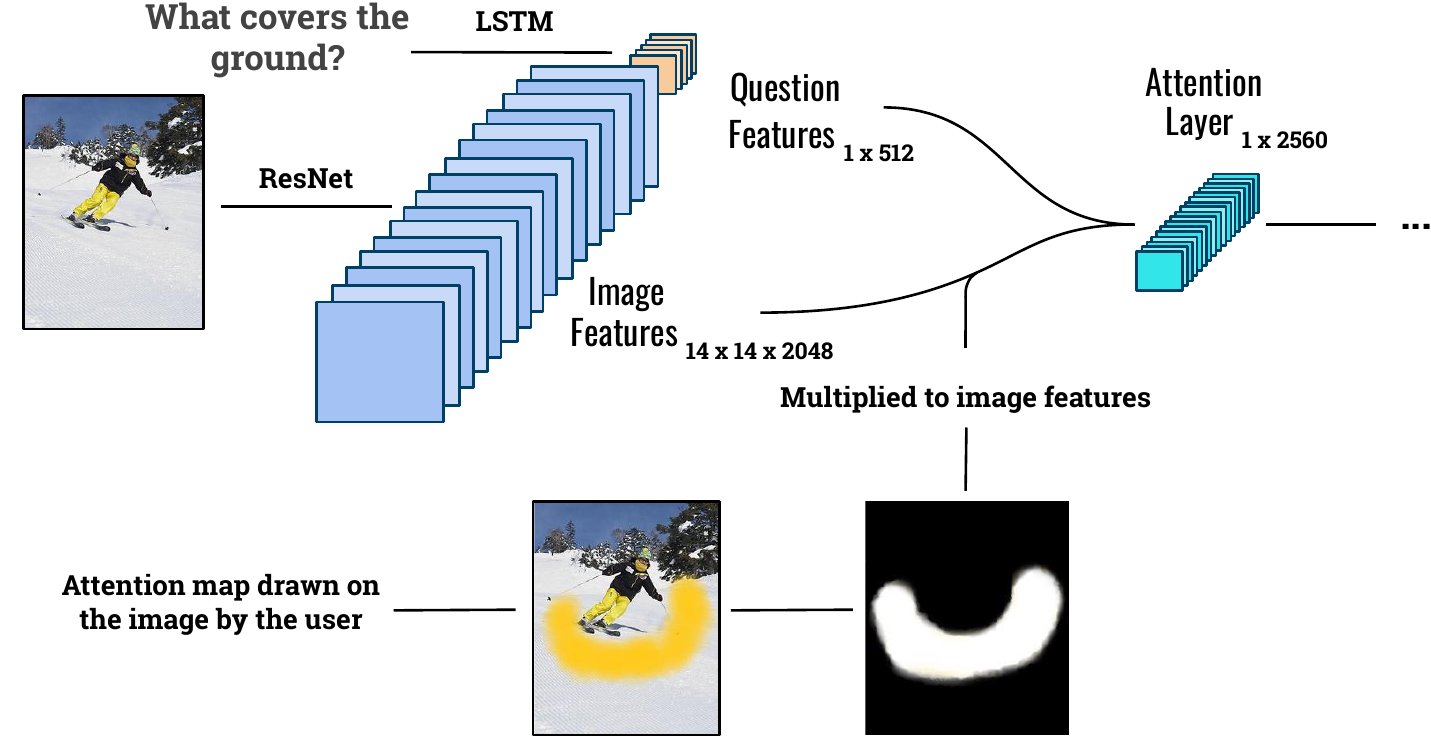}
    \caption{The architecture of active attention loop within the XVQA model.}
    \label{fig:active_attention_architecture}
\end{figure}

In the feedback loop, the model directly multiplies the user-generated attention map into the image feature map (figure \ref{fig:active_attention_architecture}). This operation accentuates the image features in the highlighted areas and mitigates the features in irrelevant sections of the image.

The purpose of this operation is to allow the subject to get involved in the inference process and provide feedback to the model interactively. In cases where the model answers the questions incorrectly, subjects attempt to correct the model's response by drawing the attention. Otherwise, for those cases where the model is already accurate, subjects try to create a different answer by altering the attention map.

\subsection{Bounding boxes}
\begin{figure}[ht]
\centering
\begin{tabular}{@{}c@{}}
  \centering
  \includegraphics[width=.45\columnwidth]{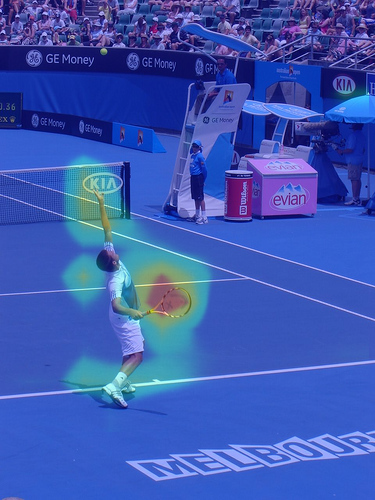}
 \end{tabular}
\begin{tabular}{@{}c@{}}
  \centering
  \includegraphics[width=.45\columnwidth]{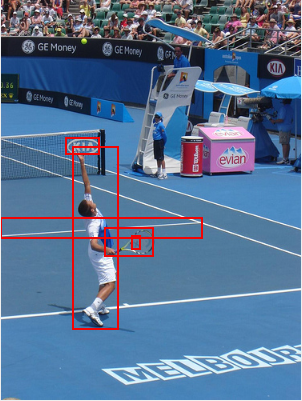}
 \end{tabular}
    \caption{Bounding box explanations generated based on spatial attention weights for the question "What is the man doing?".}
    \label{fig:bbox}
\end{figure}

The bounding boxes in this model are generated based on the annotations in the Visual Genome dataset and can carry important information about the scene. A combination of the attention maps created by the model and these annotations can produce explanations of the system behavior on a conceptual level. We calculate the average attention weight of the bounding boxes in the image based on the spatial attention maps and keep the top $K$ ($K=5$ in our studies) boxes as an indicator of most related objects in the scene contributing to the system's answer (figure \ref{fig:bbox})

\subsection{Scene graph}
\begin{figure}[ht]
\centering
  \includegraphics[width=1.0\columnwidth]{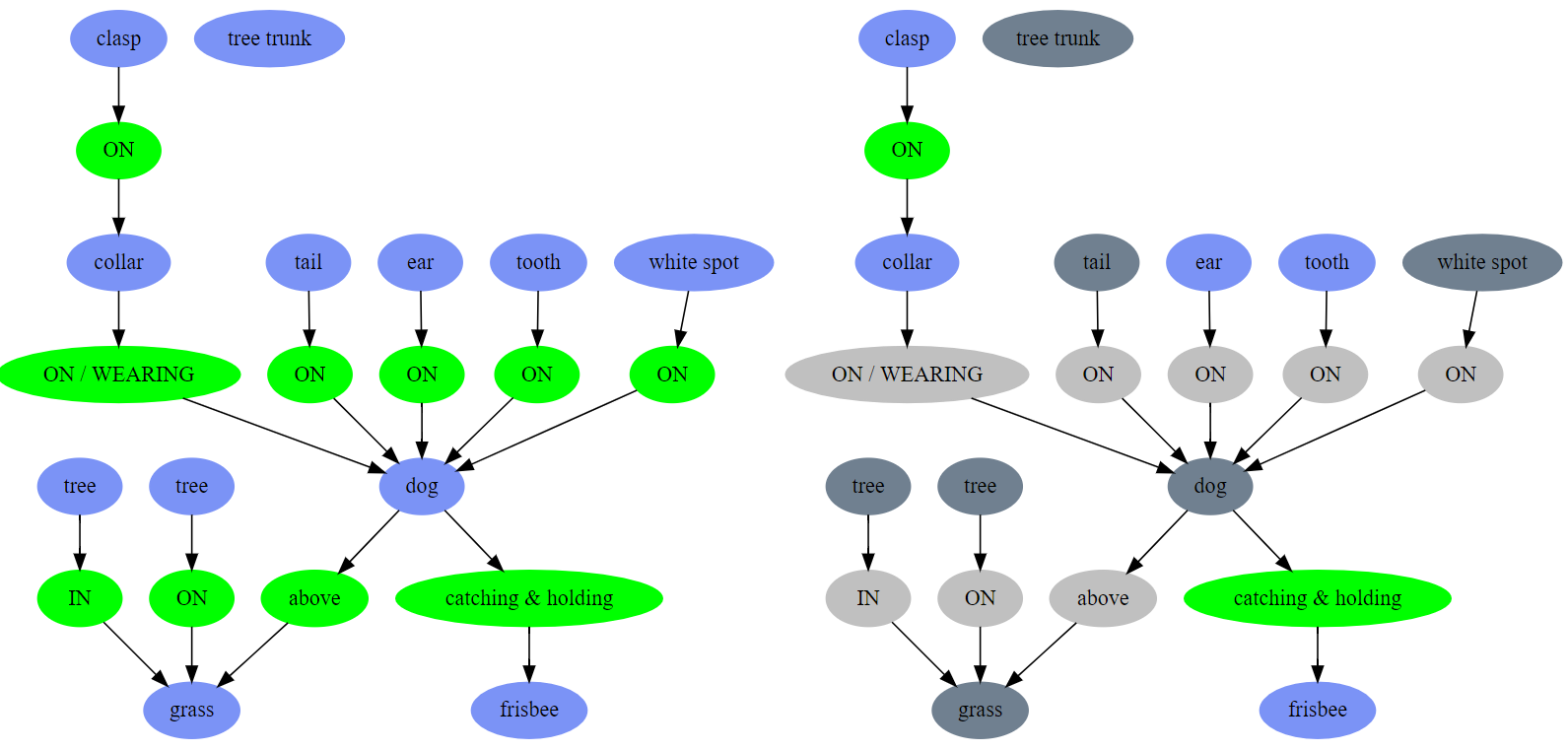}
    \caption{Left: input scene graph. Right: scene graph filtered based on the attention map weights generated by the model in response to a question.}
    \label{fig:scenegprah}
\end{figure}

The bounding box annotations are completed by the scene graph information which illustrates the relationships between different objects in the scene. The connections are in the form of \textit{subject-predicate-object} phrases and can indicate object attributes or interactions. In the Visual Genome (VG) dataset, the object labels, their bounding boxes and the scene graph connecting them provide a structured, formalized representation of components in each image \cite{Krishna2017}. For each question, we filter objects in the scene graph based on the attention weights of their bounding boxes (figure \ref{fig:scenegprah}). The users can interactively locate active objects of the scene graph and see their bounding boxes in the input image.

\subsection{Object attention}\label{sec:objatt}
Inspired by previous work \cite{ray2019lucid}, we added a MASK-RCNN image encoder to our model to produce explanations on object-level. This encoder is specifically used by the XVQA model, as the VQA model still uses Resnet encoder to produce answers.

\begin{figure}
  \centering
  \begin{subfigure}[b]{0.49\linewidth}
    \includegraphics[width=\linewidth]{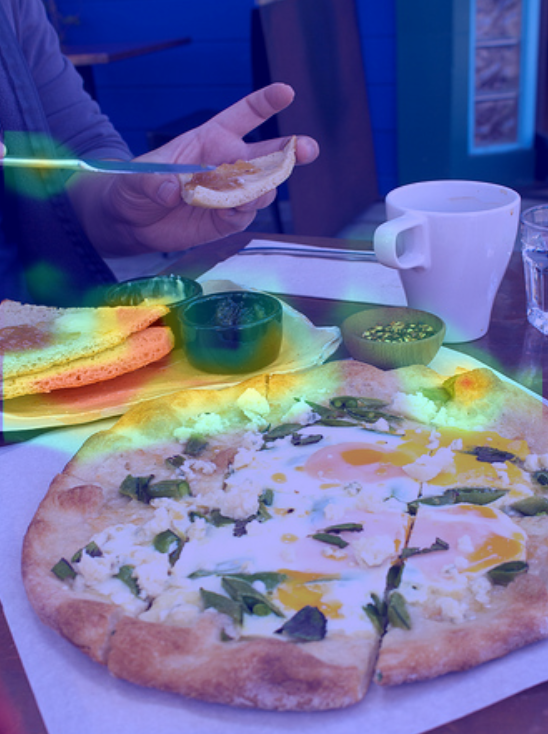}
    \caption{Spatial attention}
    \label{fig:objatt_spatt}
  \end{subfigure}
  \begin{subfigure}[b]{0.49\linewidth}
    \includegraphics[width=\linewidth]{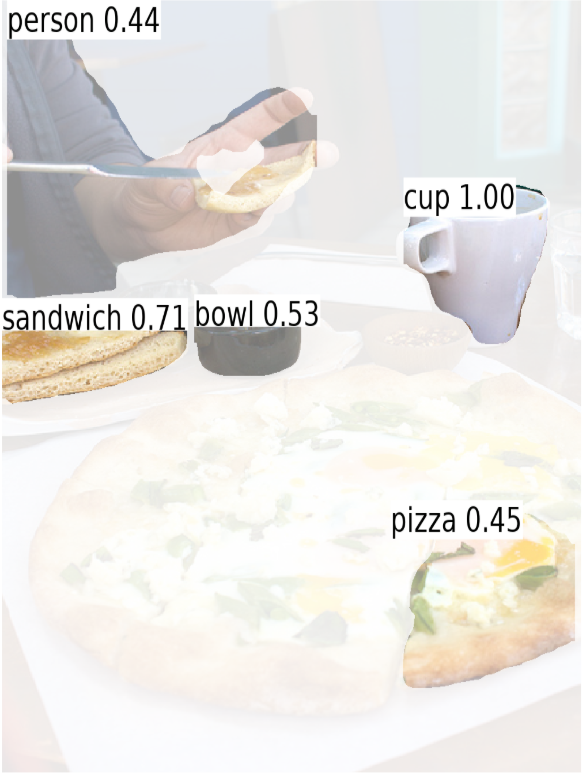}
    \caption{Object attention}
    \label{fig:objatt_objatt}
  \end{subfigure}
\caption{(b) Object-level attention compared with (a) spatial attention.}
\label{fig:objatt}
\end{figure}

Model creates object attention masks based on attention modules to highlight objects with greater contributions to the inference process. As opposed to spatial attention explanations, object attentions have the capability to segment certain entities in the scene to illustrate a more meaningful explanation for system answers(Figure \ref{fig:objatt}). For more details on the implementation of this technique, please refer to \cite{ray2019lucid}.

\subsection{Textual explanation}
Along with visual explanations, we also integrate natural language (NL) explanations in our XVQA system. Our technique is derived from the work done by \cite{ghosh2019generating} which uses the annotations of entities in an image (extracted from the scene graph), and the attention map generated by a VQA model while answering the question. 

For a given question-image pair, our textual explanation module uses the visual attention map to identify the most relevant parts of the image. The model then retrieves the bounding boxes of entities that highly overlap with these regions. 

\begin{figure}[ht]
    \centering
    \includegraphics[width=1.0\columnwidth]{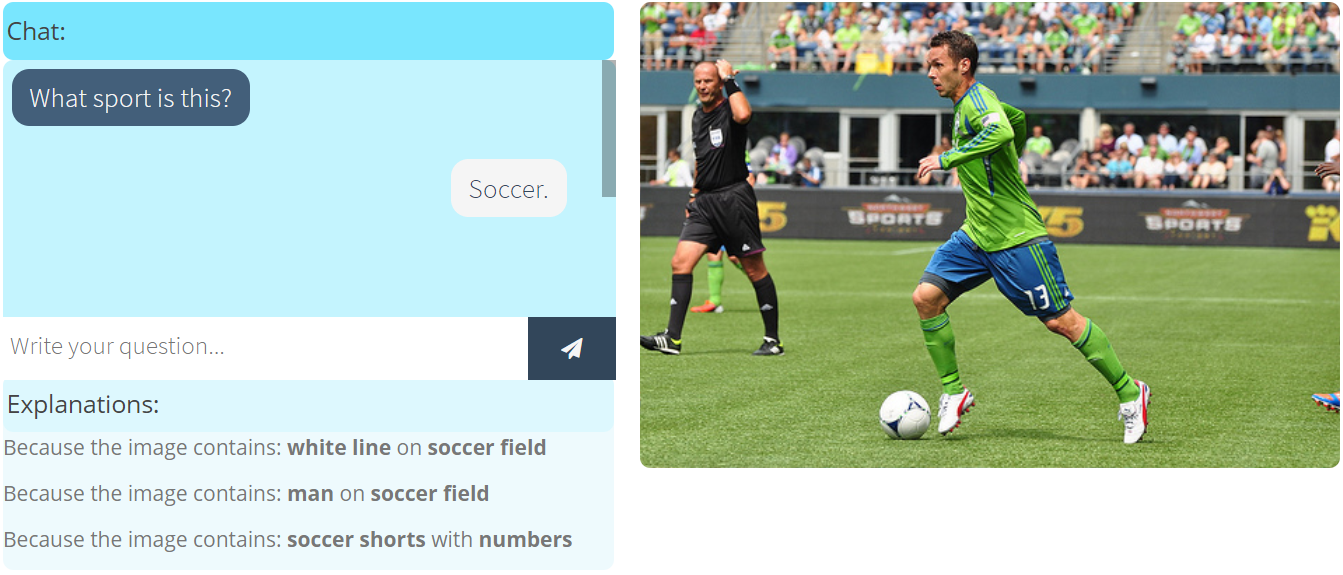}
    \caption{Sample results from the NL module producing a textual explanation for the model's answer.}
    \label{fig:nlp}
\end{figure}

The model eventually identifies those entities most relevant to the answer based on their spatial relevance on the image and their NL representation. The region descriptions for the most relevant entities form the textual explanations. A sample output generated by this technique is illustrated in figure \ref{fig:nlp}.

\section{Experimental design}
For a careful evaluation of all mentioned explanation modes, we implement an interactive interface where users can take part in a user-machine prediction task. The test starts with an introduction section and continues in the form of a series of trials where the task in each trial is to estimate the VQA system's answer correctness.

Within the introduction section, the subjects are also informed of their interaction with an AI system without any implications of its accuracy to avoid any prior bias in their mental model of the system. The subjects are also provided with a set of instructions to perform the tasks and work with the interface effectively.

\subsection{User task}
On each trial, users enter their prediction of whether they think the system would answer the system correctly or not and then declare their level of confidence in their answer on a Likert scale. Afterward, the subjects view the ground-truth, system’s top-five answers, and their probabilities in order. The system also provides its overall confidence/certainty based on normalized Shannon entropy of the answer probability distribution.

To prevent the effect of fatigue on performance in groups with longer trials, the test for each subject is limited to a one-hour session. Participants are asked to go through as many trials as possible within that period. 

\subsection{Trials}
There are two types of trials in the experiment: no-explanation trials, and explanation trials. In no-explanation trials, subjects estimates system's accuracy only based on the input image and question. 

\begin{figure}[ht]
    \centering
    \includegraphics[width=1.0\columnwidth]{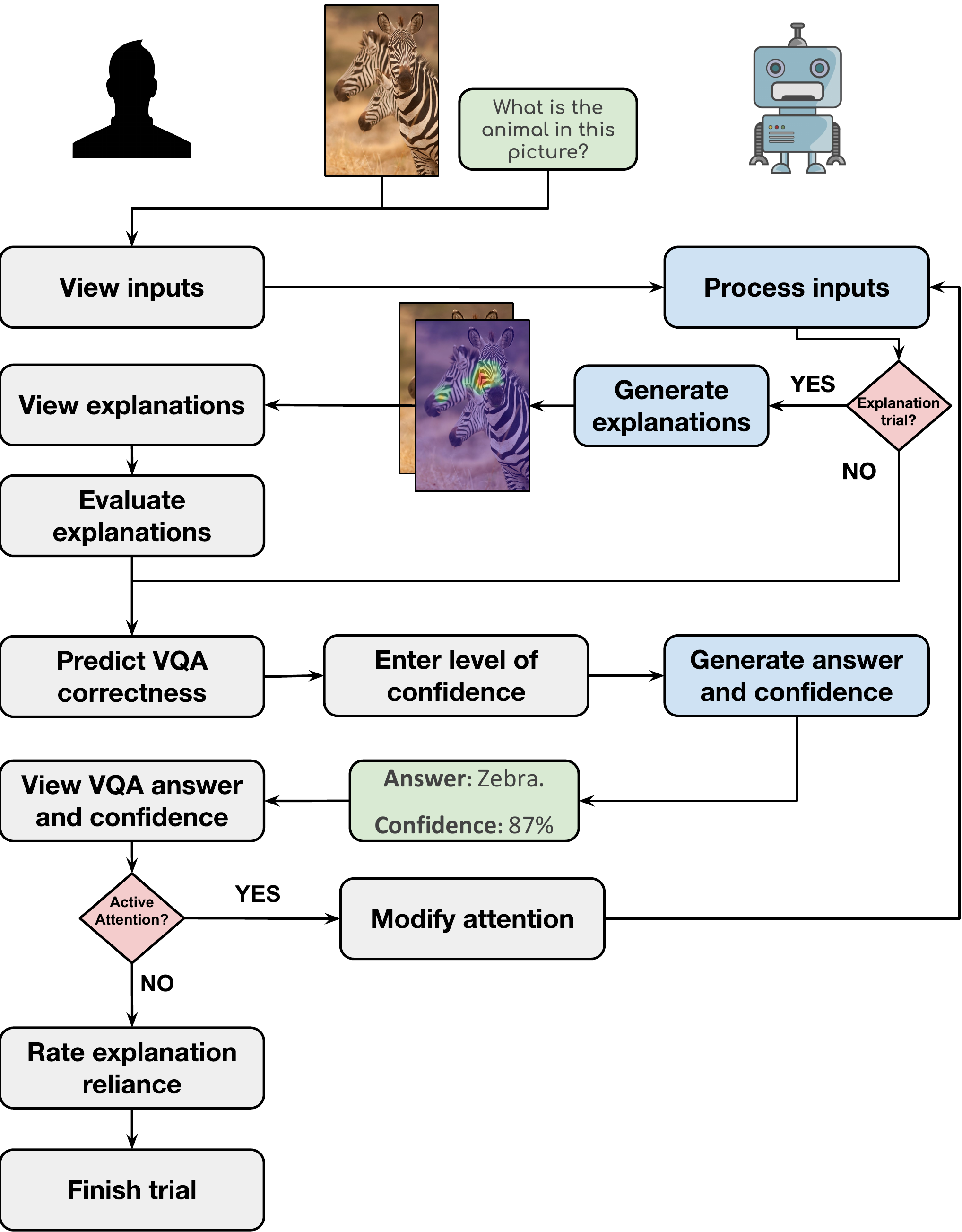}
    \caption{Flow-chart for a prediction evaluation task. The "Explanation trial?" conditional defines the type of trial as either explanation or control. The "Active attention?" conditional activates the feedback loop in case of active attention explanations.}
    \label{fig:flowchart}
\end{figure}

In explanation trials, the subjects first see the inputs and system's explanations. Before estimating the correctness of system's answer, subjects are asked to rate each explanation's helpfulness towards better predicting system's accuracy. At the end of each explanation trial, subjects rate their reliance on the explanations to predict system's accuracy. Figure \ref{fig:flowchart} depicts the order of actions in a trial in our evaluation system. 

Each test session starts with a practice block consisting of two trials. The practice trials are only purposed to familiarize the subjects with the flow of the test and are not considered in any of the final results. The rest of the test is carried out in blocks where each block includes five trials.

\subsection{Study groups}
The study involves six groups of participants. The control group (NE) does not see any explanation modes, so its task is reduced to predicting the system's correctness in trials. The explanation groups are exposed to either one or a combination of explanation modes before they make their prediction about the system's answer.

\begin{figure}[ht]
    \centering
    \includegraphics[width=1.0\columnwidth]{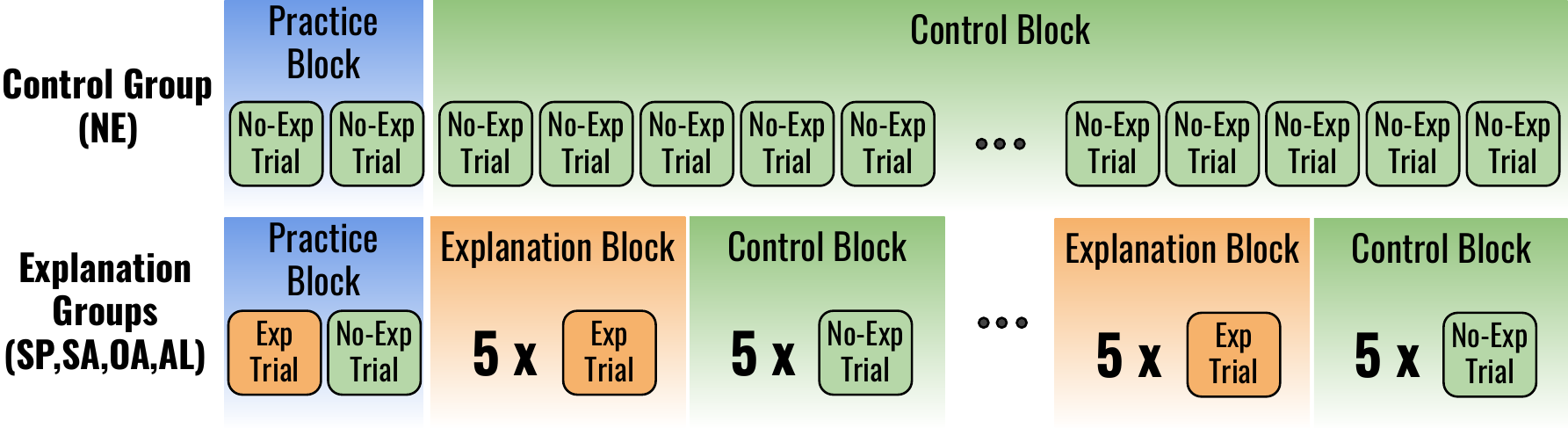}
    \caption{Structure of the test sessions in control group (NE) and explanation groups.}
    \label{fig:test_design}
\end{figure}

The control group (NE) only sees a block of no-explanation trials throughout the whole test. For the groups with explanation modes, the blocks toggle between explanation and no-explanation modes. The no-explanation blocks in explanation groups act as control tests to assess prediction quality and mental model progress as the users see more trials. (figure \ref{fig:test_design})   

The explanation blocks view the explanations generated by the model before the users make their predictions and show the answer from the system along with the system's confidence afterwards. The no-explanation blocks only ask for user's prediction without exposing any explanations beforehand.  

Group SA has an interactive workflow within which subjects first go through to the spatial attention explanation and then modify the attentions in a feedback loop. Each explanation group is dedicated to a specific explanation mode, except group SE which combines bounding box, scene graph and textual explanations. The study was conducted with 90 participants and a total number of more than 10,000 trials. Table \ref{table:teststats} shows the number of participants in each group and the number of trials in each group.

\begin{table}[ht]
\caption{User Study Design and Statistics.}\smallskip
\centering
\resizebox{.95\columnwidth}{!}{
\smallskip\begin{tabular}{|c|l|c|c|}
\hline
\multicolumn{2}{|c|}{Group} & {Subjects} & {Trials} \\
\hline
NE & Control group & 15 & 4124 \\
SP & Spatial attention & 15 & 1826 \\
SA & Active attention & 15 & 1021 \\
SE & Semantic & 15 & 1261 \\
OA & Object attention & 15 & 1435 \\
AL & All explanations & 15 & 846\\
\hline
 \multicolumn{1}{r}{} & \multicolumn{1}{r|}{Total} & 90 & 10,513\\
 \cline{3-4}
\end{tabular}
}
\label{table:teststats}
\end{table}

A total number of 3969 image-question pairs were randomly selected from the overlap of VG dataset \cite{Krishna2017} and VQA dataset \cite{balanced_vqa_v2} to be used in the trials. The questions asked on each trial is selected from the VQA dataset and the annotations used in generating the explanations are extracted from the VG dataset. In the selection, all yes-no and counting questions were excluded to draw the focus of the test to non-trivial questions and less obvious answers with higher levels of detail in explanations.

\section{Results}

After assigning different groups of participants to specific combinations of explanations (including a control group that received no explanations) and having them perform the VQA prediction task, we evaluated different hypotheses about the explanations' impact on various aspects of human-machine task performance. The results are compared either based on the average of all trials within certain groups or based on the progress throughout the tests. Since the task in each group and trial can be different than other groups and trials, the number of trials finished by subjects vary between groups and even within groups.
\subsection{Impact on user-machine task performance} 

\begin{figure}[ht]
    \centering
    \includegraphics[width=1.0\linewidth]{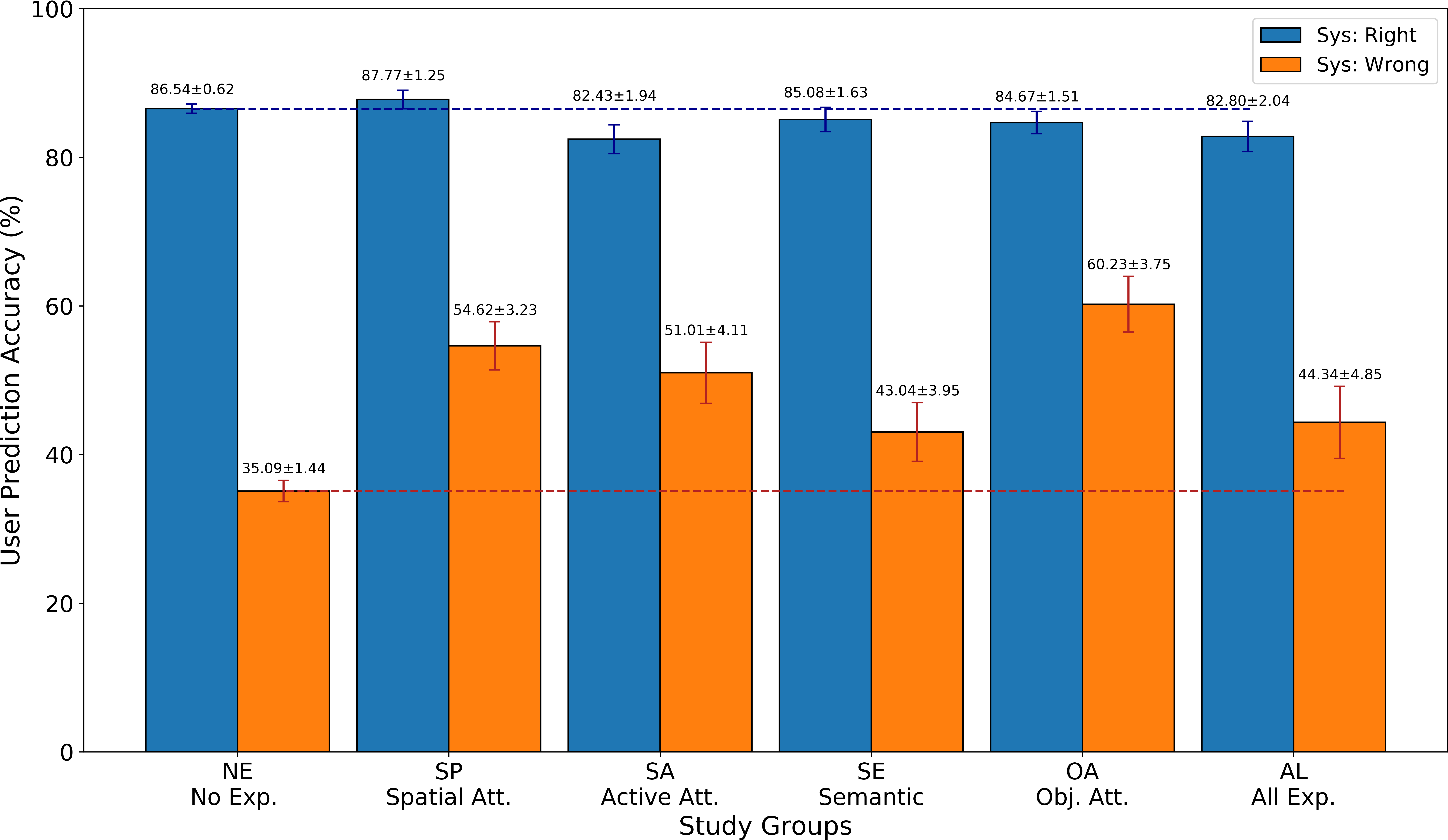}
    \caption{The average values of user's prediction accuracy (user performance) compared between different groups (sys:right p $=$ 0.061, sys:wrong  p$<$ 0.0001, overall p$=$0.0001).}
    \label{fig:acccomp_all}
\end{figure}

The first metric we used to assess user-machine task performance is the user's accuracy for predicting the machine's correctness, and whether this is affected by explanations. We tested for any effect (positive or negative) between accuracy and presence of explanations, using a chi-squared test. The results from different groups show an overall accuracy increase in all explanation groups compared to the control group, however this is statistically significant only for  cases where the system's answer is wrong (see figure \ref{fig:acccomp_all}).

\begin{figure}
  \centering
  \begin{subfigure}[b]{0.495\linewidth}
    \includegraphics[width=\linewidth]{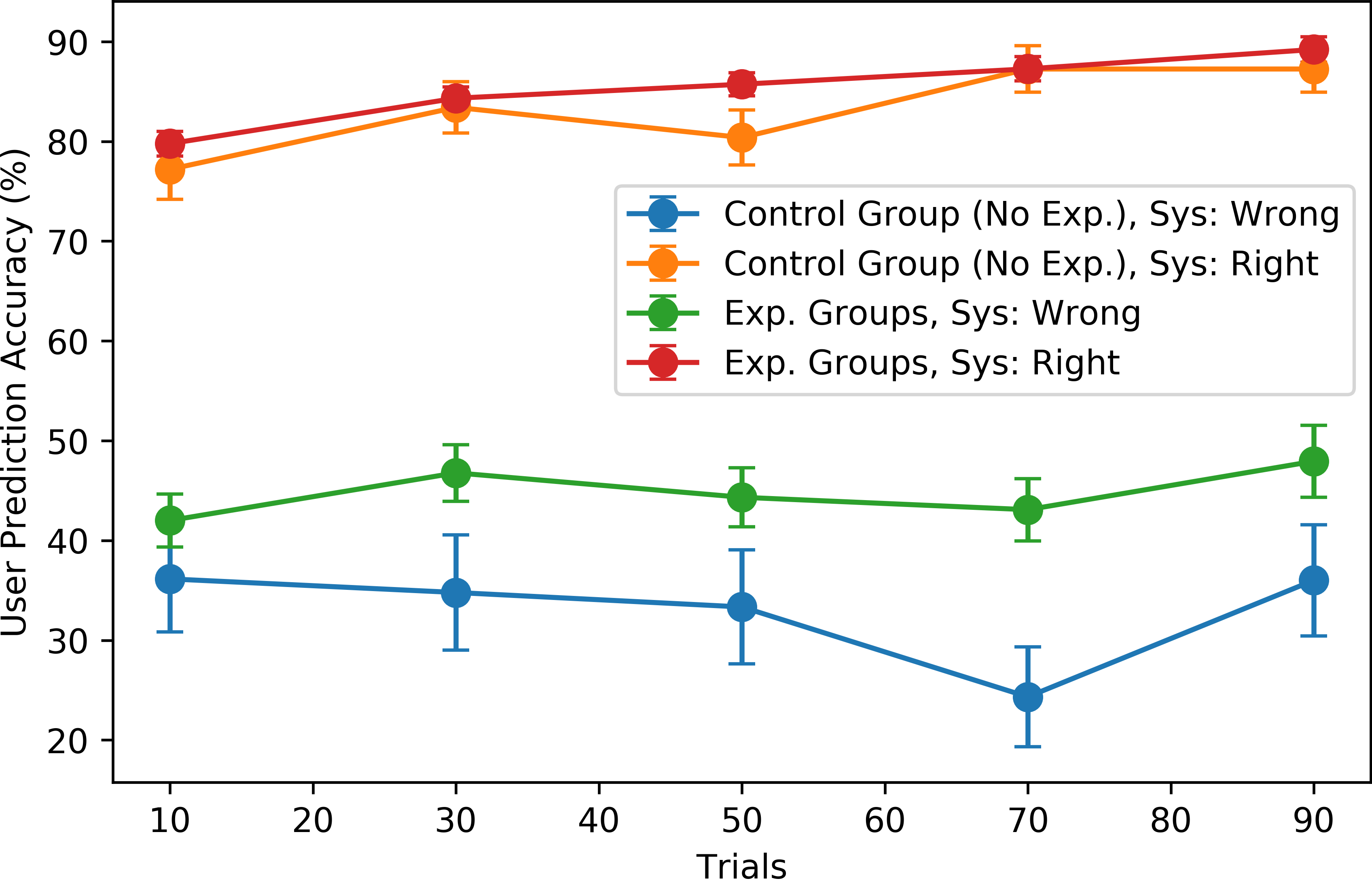}
    \caption{}
    \label{fig:accProg_comp_ctrl_exp}
  \end{subfigure}
  \begin{subfigure}[b]{0.495\linewidth}
    \includegraphics[width=\linewidth]{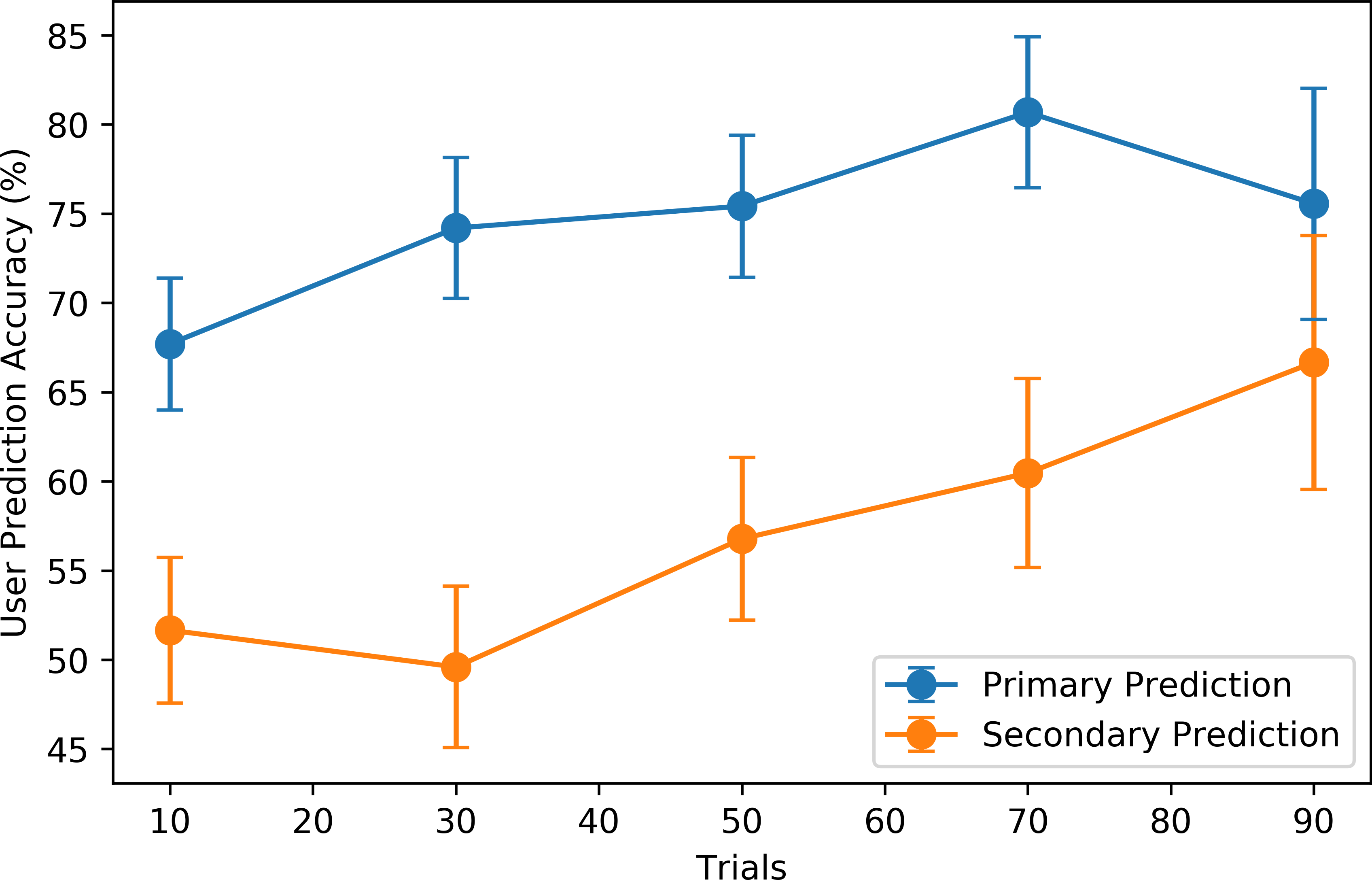}
    \caption{}
    \label{fig:actAtt_accComp_feedback}
  \end{subfigure}
\caption{(a) The progress of user's prediction accuracy compared between the control group and all explanation groups. The results are separated based on the accuracy of the system. (b) Prediction accuracy progress in Active Attention explanation group. primary prediction is made based on model's original attention map and secondary prediction is based on the modified attention map that the subject provides.}
\label{fig:accProg_accComp}
\end{figure}

The progress of prediction accuracy is also another metric to quantify subject's progress in understanding and predicting systems behaviour. As subjects go through trials in different groups, we compare the improvement of their mental model based on their prediction accuracy (figure \ref{fig:accProg_comp_ctrl_exp}). As shown in figure \ref{fig:accProg_comp_ctrl_exp}, in both cases whether the system is right or wrong, the subjects in explanation groups show a more steady improvement in their prediction accuracy. 

\subsection{User explanation helpfulness ratings} 

\begin{figure}
  \centering
  \begin{subfigure}[b]{0.495\linewidth}
    \includegraphics[width=\linewidth]{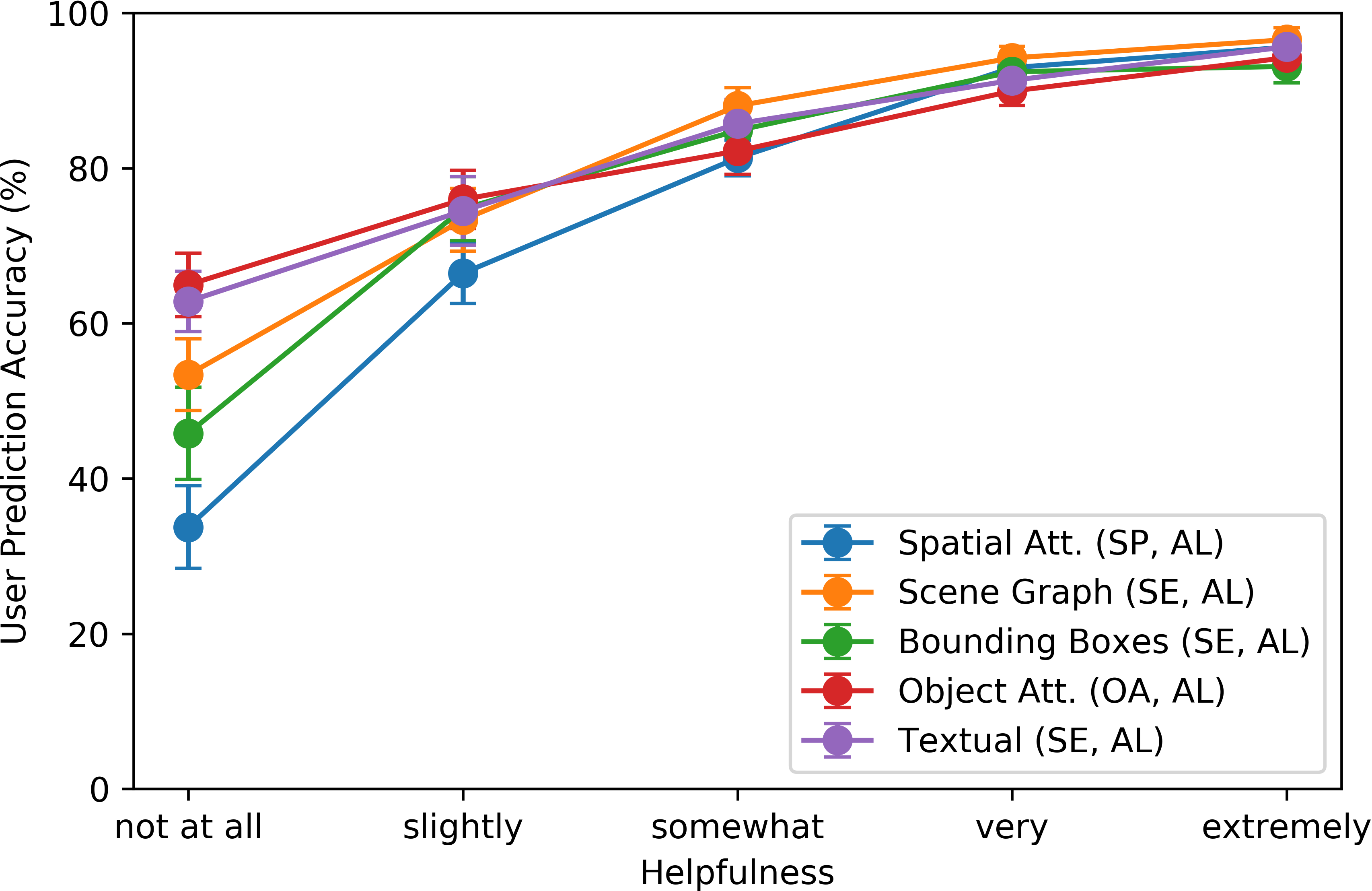}
    \caption{}
    \label{fig:acchlp_wrong}
  \end{subfigure}
  \begin{subfigure}[b]{0.495\linewidth}
    \includegraphics[width=\linewidth]{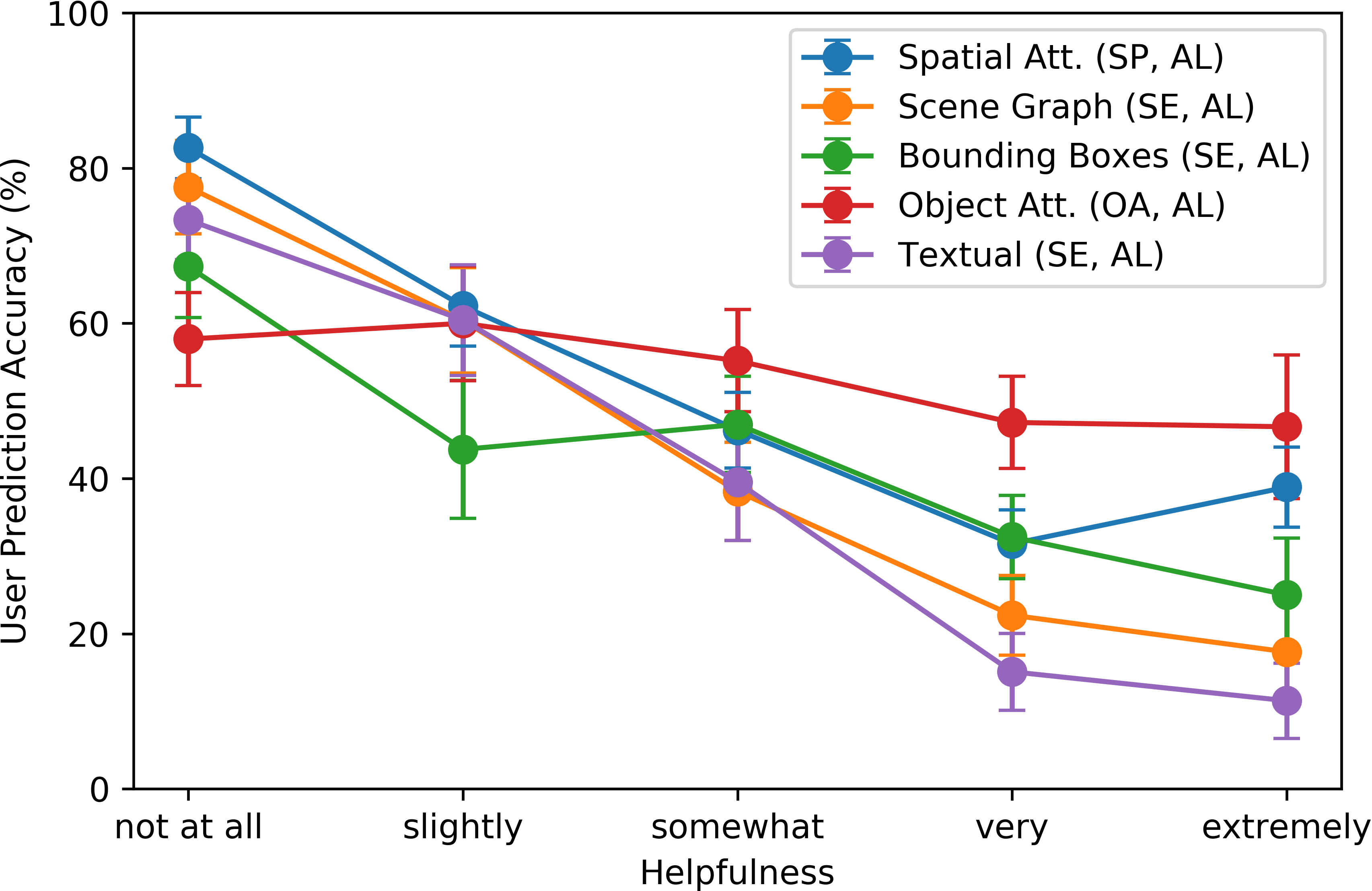}
    \caption{}
    \label{fig:acchlp_right}
  \end{subfigure}
\caption{The average values of user's prediction accuracy for each explanation mode vs. user's ratings on explanations' helpfulness for cases where (a) the system is right, and (b) where the system is wrong. (p$<$ 0.0001)}
\label{fig:acchlp_wrong_right}
\end{figure}

 Before making a prediction about the VQA model's answer, users rate each explanation mode based on how much it helped them in the prediction task, a rating that we call "explanation helpfulness". Comparing these helpfulness ratings with the users' prediction accuracy reveals a positive correlation with accuracy improvement (accuracy after minus accuracy before) and helpfulness of explanations, but only in cases where the system is \emph{right}. Figure \ref{fig:acchlp_right} implies that when users find explanations helpful, they do better on the prediction task. On the other hand, a higher rating for explanations when the system is wrong (figure \ref{fig:acchlp_wrong}) has lead to lower human prediction accuracy. This observation shows the effective role of explanations in the process of decision making for users.
 
\subsection{Active Attention explanation}

Within group SA, subjects view and interact with active attention explanations before making their prediction. Similar to spatial attention, users first make a prediction based on the attention map made by VQA model. On the second step, subjects draw a new attention map for the model in the purpose of changing networks answer. Subjects can compare their attention with model's attention and the answer created based on each of them. Figure \ref{fig:actAtt_accComp_feedback} illustrates the trend of prediction accuracy progress as subjects interact with active attentions. While the explanation helps subjects improve their primary prediction of system's correctness, they also substantially improve in predicting system when working with their modified attention (secondary prediction).

\subsection{Impact of Active Attention on user confidence}

Active attention explanation provides users with a feedback loop to modify the system's attention and see the result of attention changes in the model's answer. In trials with active attention explanation, users make two predictions: one based on the original spatial attention provided by the user, and a secondary prediction after they modify the attention map. We consider the accuracy of the primary prediction as an indicator of the user's mental model state. The secondary prediction is more specifically dependant on users general mental model of the attention map.

\begin{figure}
  \centering
  \begin{subfigure}[b]{0.495\linewidth}
    \includegraphics[width=\linewidth]{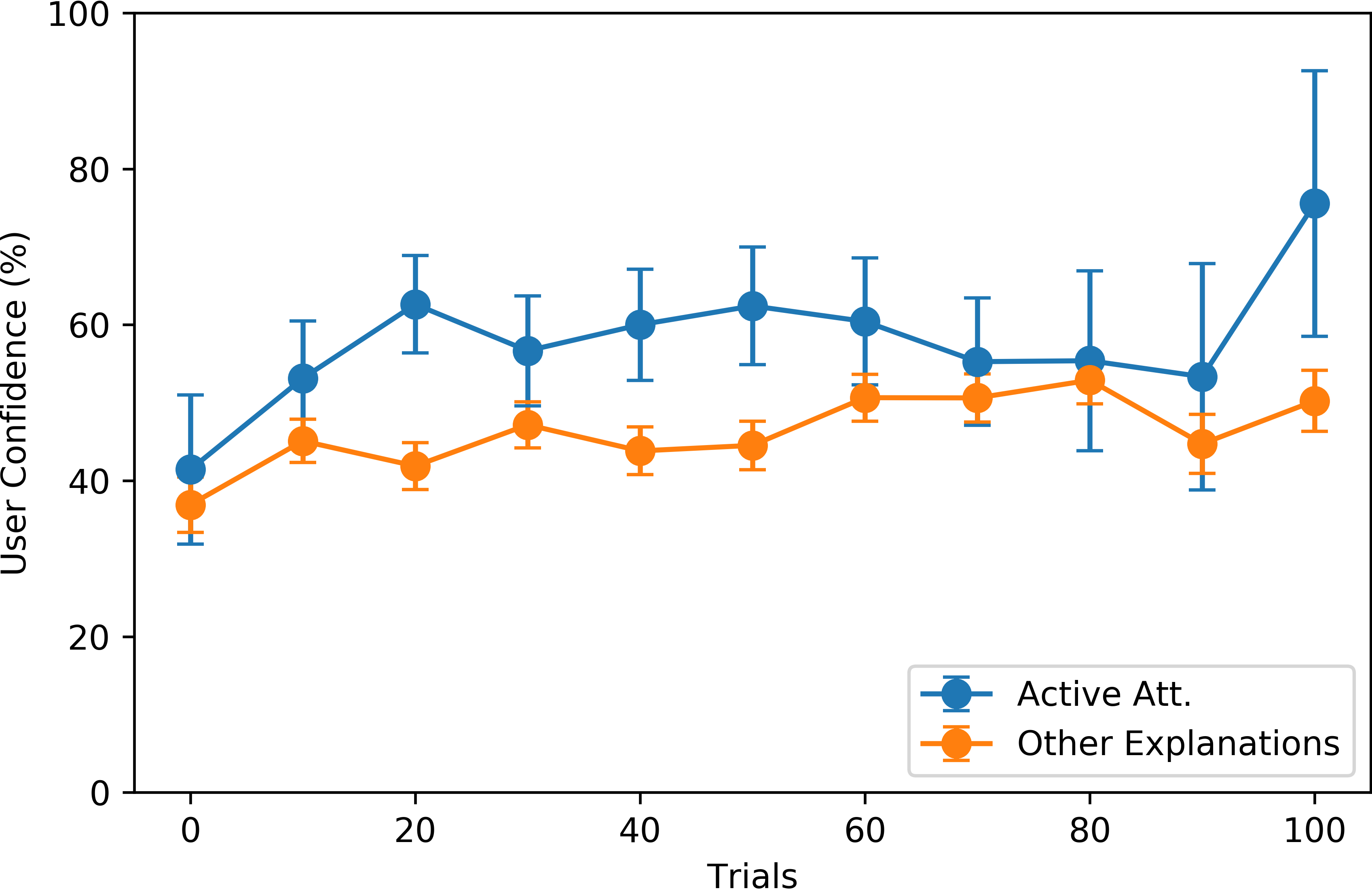}
    \caption{}
    \label{fig:confProgActAtt}
  \end{subfigure}
  \begin{subfigure}[b]{0.495\linewidth}
    \includegraphics[width=\linewidth]{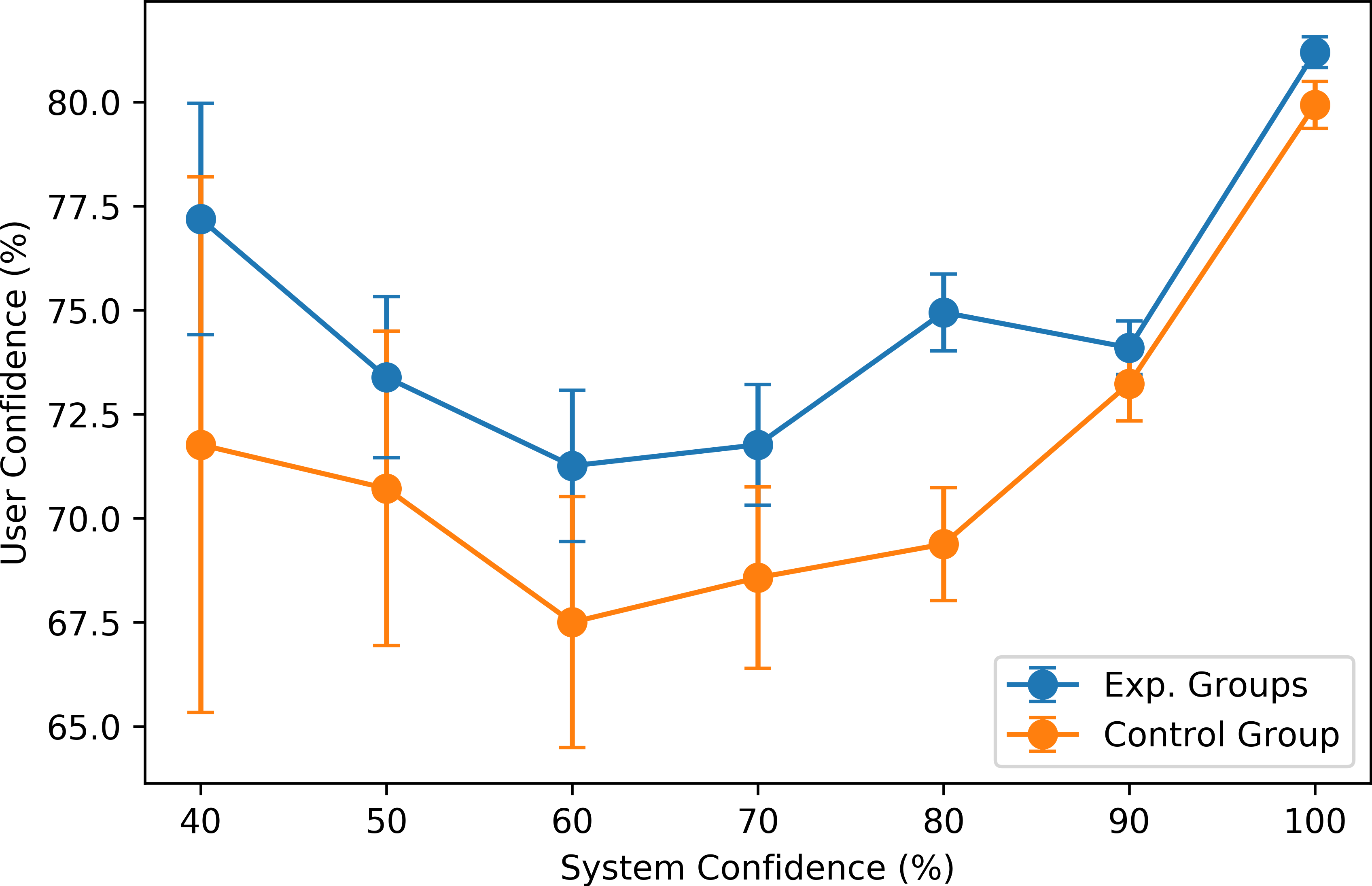}
    \caption{}
    \label{fig:sysconfusrconf}
  \end{subfigure}
\caption{(a) User confidence progression comparison between the active attention group and other groups. (b) User confidence in prediction vs. system confidence in answer.}
\label{fig:user_confidence_plots}
\end{figure}

Comparing results from different explanation groups with the active attention group shows that users in the active attention group have higher average confidence in their primary predictions compared to other explanation groups (see figure \ref{fig:confProgActAtt}). 

While the increase in user confidence points out the confidence and trust built by the active attention explanation, the average prediction accuracy in this group of participants is lower than other groups. These results suggest a higher potential for this technique to produce real insight into the model if used in multiple feedback loops instead of just one.

\subsection{Impact on trust and reliance} 

 Another important purpose of explanation systems is to create user trust in AI machines so that they can rely on the outcome of the system.
 
 In our user study, we ask users about their level of reliance (in Likert scale) on the explanation section while predicting system's performance. Comparing users reliance with respect to their performance indicates a correlation between the reliance and users accuracy in those cases when the system is wrong (Figure \ref{fig:accVsRel}).
 
 Moreover, users declare their level of confidence in their prediction on a Likert scale. Generally, we can assume the users' level of confidence in their prediction as a function of user confidence in the system and also the system's confidence in its answer. In the control group with no explanations, the level of confidence mainly stems from system performance in previous trials (mental model); while in other groups, the explanations have a direct effect on the level of confidence.
 
 Figure \ref{fig:sysconfusrconf} shows average user confidence compared with system confidence (provided to users after they make their predictions) in those cases when user's prediction is correct. The results indicate a consistent increase in user confidence when exposed to explanations against the control group with no explanations.
 
\subsection{Impact of explanation goodness} 

\begin{figure}
  \centering
  \begin{subfigure}[b]{0.495\linewidth}
    \includegraphics[width=\linewidth]{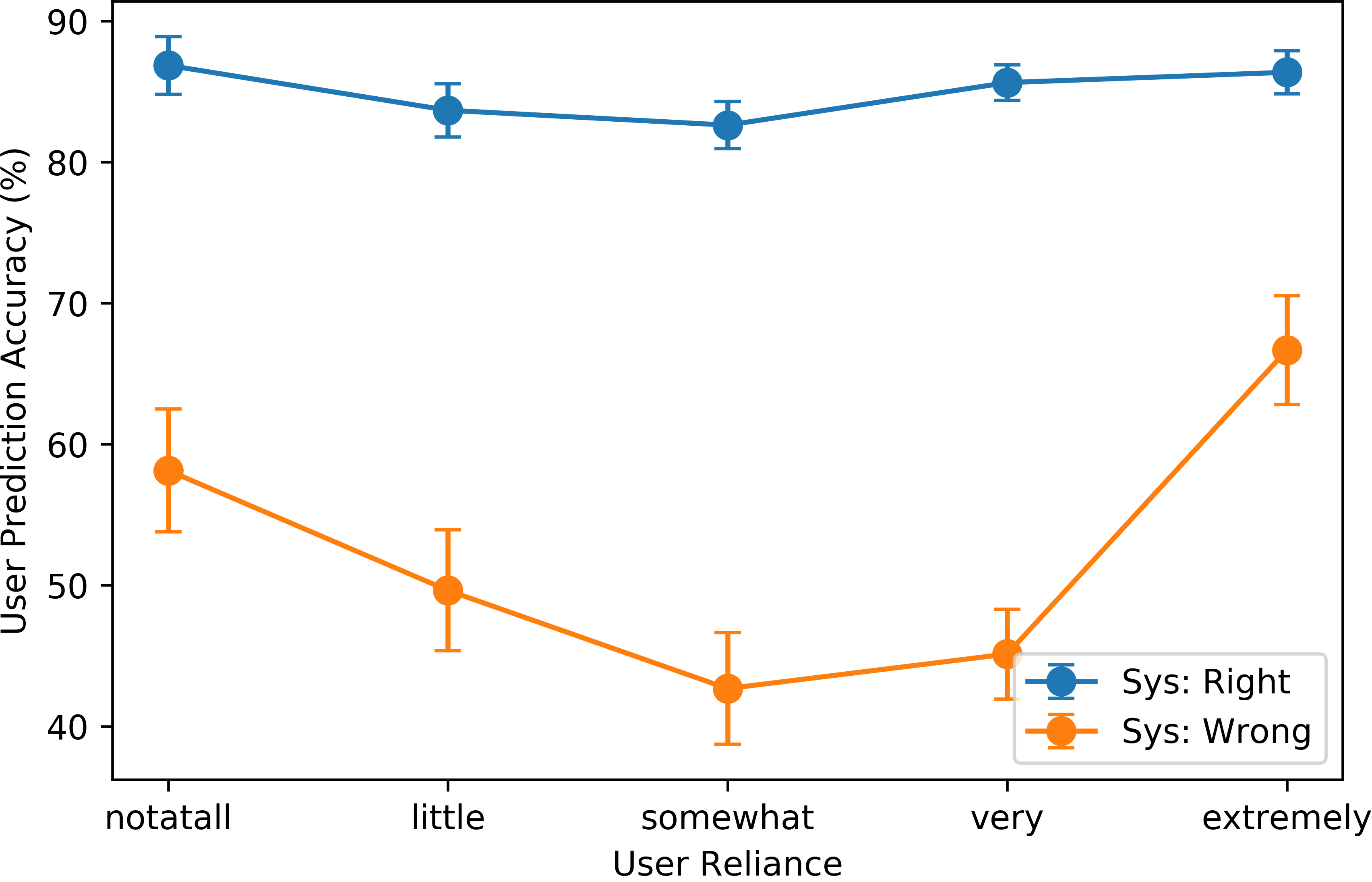}
    \caption{}
    \label{fig:accVsRel}
  \end{subfigure}
  \begin{subfigure}[b]{0.495\linewidth}
    \includegraphics[width=\linewidth]{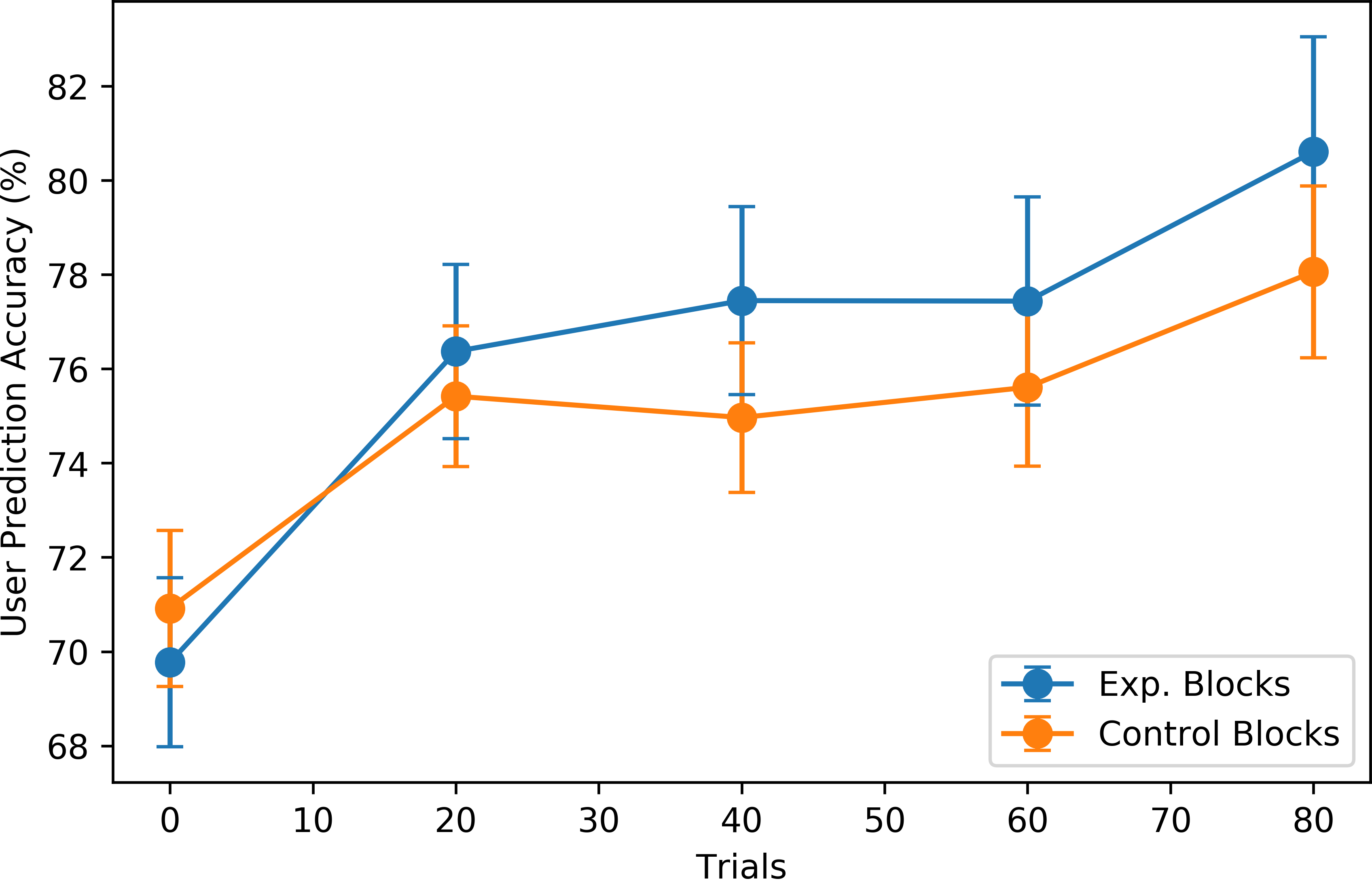}
    \caption{}
    \label{fig:acccomp_exp_noexp}
  \end{subfigure}
\caption{(a) Users prediction accuracy vs. their reliance on explanation divided by the accuracy of system. (b) Prediction accuracy growth in explanation groups compared between exp. blocks and no-exp. blocks.}
\label{fig:accVsRel_accComp}
\end{figure}

As mentioned earlier, in explanation groups users go through blocks of trials. To assess the goodness of explanations in helping users predict systems answer, each block of trials with explanation is followed by a block without explanations. Comparing the user prediction accuracy between these blocks illustrates the progress of users mental model in presence of explanations (Figure \ref{fig:acccomp_exp_noexp}). Results indicate that within explanation blocks users have built a better mental model to predict system and made progress in understanding system answers.

\section{Discussion}

The overall assessment of user performance reveals a substantial improvement of prediction accuracy in the presence of explanations while the system is not correct. Users also provide higher ratings for the explanations when they perform better and vice versa. This direct correlation in all explanation modes strongly suggests the effectiveness of these explanations within the prediction task.

In group AL, although the subjects viewed all explanation modes, yet we do not see a higher level of accuracy compared to other groups. The feedback from the post-study interviews pointed out two possible reasons for such observation:
1) the overwhelming amount of information in the group decreased the performance level for the subjects; 2) those cases where explanation modes conflicted with each other confused some of the subjects.

Users show higher levels of confidence when exposed to active attention in explanation groups; although, the overall performance of the active attention group (SA) does not yet exceed the spatial attention group (SP). The reason behind this drawback can be active attention's limit to only visual features and not question features. Possibly, multiple feedback loops can also help users better understand the role of image features as only one of (and not all of) the contributors in the final answer.

In cases where the system is wrong, user's accuracy show an interesting correlation with user's reliance. The subjects seem to do well either when they are extremely relying on the explanations or when they are completely ignoring them. For those cases that the users ignore the explanations, post-study interviews suggest that the subjects made their decision based on their mental model of the system and previous similar trials.

\section{Conclusion}

We designed an interactive experiment to probe explanation effectiveness in terms of improving user prediction accuracy, confidence, and reliance in the context of a VQA task. The results of our study show that the explanations help to improve VQA accuracy, and explanation ratings approve the effectiveness of explanations in human-machine AI collaboration tasks.

To evaluate various modes of explanations, we conducted a user study with 90 participants. Users interactively rated different explanation modes and used them for predicting AI system behavior. The user-machine task performance results indicate improvements when users were exposed to the explanations. User confidence in predictions also improved when they viewed explanations which display the potential of our multi-modal explanation system in building user trust.

The strong correlation between the users' rating on explanation helpfulness and their performance in the prediction tasks shows the effectiveness of explanations in the user-machine task performance. Those explanations identified as more helpful helped users in cases where the system was accurate. On the other hand, in cases where the system was inaccurate, those explanations ranked as more helpful became more misleading.

We also introduced an interactive explanation mode (active attention) where users could directly alter the system's attention and receive feedback from it. Comparing the user confidence growth between active attention and other explanation groups shows a higher level of trust built in users, which shows the effectiveness of interactive explanations in building a better mental model of the AI system.

As a future direction, we may investigate other interactive explanation modes to maximize the performance in human-machine tasks. On the other hand, user feedback and ratings for the different modes explored in this study can guide us towards more effective explanation models in XAI systems.

\section{Acknowledgments}

This research was developed with funding from the Defense Advanced Research Projects Agency (DARPA) under the Explainable AI (XAI) program. The views, opinions and/or findings expressed are those of the author and should not be interpreted as representing the official views or policies of the Department of Defense or the U.S. Government.

\bibliography{egbib}

\begin{thebibliography}{}

\bibitem[\protect\citeauthoryear{Anne~Hendricks \bgroup et al\mbox.\egroup
  }{2018}]{anne2018grounding}
Anne~Hendricks, L.; Hu, R.; Darrell, T.; and Akata, Z.
\newblock 2018.
\newblock Grounding visual explanations.
\newblock In {\em Proceedings of the European Conference on Computer Vision
  (ECCV)},  264--279.

\bibitem[\protect\citeauthoryear{Antol \bgroup et al\mbox.\egroup
  }{2015}]{antol2015vqa}
Antol, S.; Agrawal, A.; Lu, J.; Mitchell, M.; Batra, D.; Lawrence~Zitnick, C.;
  and Parikh, D.
\newblock 2015.
\newblock Vqa: Visual question answering.
\newblock In {\em Proceedings of the IEEE international conference on computer
  vision},  2425--2433.

\bibitem[\protect\citeauthoryear{Chandrasekaran \bgroup et al\mbox.\egroup
  }{2017}]{chandrasekaran2017takes}
Chandrasekaran, A.; Yadav, D.; Chattopadhyay, P.; Prabhu, V.; and Parikh, D.
\newblock 2017.
\newblock It takes two to tango: Towards theory of ai's mind.
\newblock {\em arXiv preprint arXiv:1704.00717}.

\bibitem[\protect\citeauthoryear{Chandrasekaran \bgroup et al\mbox.\egroup
  }{2018}]{chandrasekaran2018explanations}
Chandrasekaran, A.; Prabhu, V.; Yadav, D.; Chattopadhyay, P.; and Parikh, D.
\newblock 2018.
\newblock Do explanations make vqa models more predictable to a human?
\newblock {\em arXiv preprint arXiv:1810.12366}.

\bibitem[\protect\citeauthoryear{Cosley \bgroup et al\mbox.\egroup
  }{2003}]{cosley2003seeing}
Cosley, D.; Lam, S.~K.; Albert, I.; Konstan, J.~A.; and Riedl, J.
\newblock 2003.
\newblock Is seeing believing?: how recommender system interfaces affect users'
  opinions.
\newblock In {\em Proceedings of the SIGCHI conference on Human factors in
  computing systems},  585--592.
\newblock ACM.

\bibitem[\protect\citeauthoryear{Das \bgroup et al\mbox.\egroup
  }{2017}]{das2017human}
Das, A.; Agrawal, H.; Zitnick, L.; Parikh, D.; and Batra, D.
\newblock 2017.
\newblock Human attention in visual question answering: Do humans and deep
  networks look at the same regions?
\newblock {\em Computer Vision and Image Understanding} 163:90--100.

\bibitem[\protect\citeauthoryear{Fukui \bgroup et al\mbox.\egroup
  }{2016}]{fukui2016multimodal}
Fukui, A.; Park, D.~H.; Yang, D.; Rohrbach, A.; Darrell, T.; and Rohrbach, M.
\newblock 2016.
\newblock Multimodal compact bilinear pooling for visual question answering and
  visual grounding.
\newblock {\em arXiv preprint arXiv:1606.01847}.

\bibitem[\protect\citeauthoryear{Ghosh \bgroup et al\mbox.\egroup
  }{2019}]{ghosh2019generating}
Ghosh, S.; Burachas, G.; Ray, A.; and Ziskind, A.
\newblock 2019.
\newblock Generating natural language explanations for visual question
  answering using scene graphs and visual attention.
\newblock {\em arXiv preprint arXiv:1902.05715}.

\bibitem[\protect\citeauthoryear{Goyal \bgroup et al\mbox.\egroup
  }{2017}]{balanced_vqa_v2}
Goyal, Y.; Khot, T.; Summers{-}Stay, D.; Batra, D.; and Parikh, D.
\newblock 2017.
\newblock Making the {V} in {VQA} matter: Elevating the role of image
  understanding in {V}isual {Q}uestion {A}nswering.
\newblock In {\em Conference on Computer Vision and Pattern Recognition
  (CVPR)}.

\bibitem[\protect\citeauthoryear{He \bgroup et al\mbox.\egroup
  }{2017}]{He_2017_ICCV}
He, K.; Gkioxari, G.; Dollar, P.; and Girshick, R.
\newblock 2017.
\newblock Mask r-cnn.
\newblock In {\em The IEEE International Conference on Computer Vision (ICCV)}.

\bibitem[\protect\citeauthoryear{Hendricks \bgroup et al\mbox.\egroup
  }{2016}]{hendricks2016generating}
Hendricks, L.~A.; Akata, Z.; Rohrbach, M.; Donahue, J.; Schiele, B.; and
  Darrell, T.
\newblock 2016.
\newblock Generating visual explanations.
\newblock In {\em European Conference on Computer Vision},  3--19.
\newblock Springer.

\bibitem[\protect\citeauthoryear{Huk~Park \bgroup et al\mbox.\egroup
  }{2018}]{huk2018multimodal}
Huk~Park, D.; Anne~Hendricks, L.; Akata, Z.; Rohrbach, A.; Schiele, B.;
  Darrell, T.; and Rohrbach, M.
\newblock 2018.
\newblock Multimodal explanations: Justifying decisions and pointing to the
  evidence.
\newblock In {\em Proceedings of the IEEE Conference on Computer Vision and
  Pattern Recognition},  8779--8788.

\bibitem[\protect\citeauthoryear{Jiang \bgroup et al\mbox.\egroup
  }{2017}]{jiang2017learning}
Jiang, Z.; Wang, Y.; Davis, L.; Andrews, W.; and Rozgic, V.
\newblock 2017.
\newblock Learning discriminative features via label consistent neural network.
\newblock In {\em 2017 IEEE Winter Conference on Applications of Computer
  Vision (WACV)},  207--216.
\newblock IEEE.

\bibitem[\protect\citeauthoryear{Jiang \bgroup et al\mbox.\egroup
  }{2018a}]{jiang2018pythia}
Jiang, Y.; Natarajan, V.; Chen, X.; Rohrbach, M.; Batra, D.; and Parikh, D.
\newblock 2018a.
\newblock Pythia v0. 1: the winning entry to the vqa challenge 2018.
\newblock {\em arXiv preprint arXiv:1807.09956}.

\bibitem[\protect\citeauthoryear{Jiang \bgroup et al\mbox.\egroup
  }{2018b}]{DBLP:journals/corr/abs-1807-09956}
Jiang, Y.; Natarajan, V.; Chen, X.; Rohrbach, M.; Batra, D.; and Parikh, D.
\newblock 2018b.
\newblock Pythia v0.1: the winning entry to the {VQA} challenge 2018.
\newblock {\em CoRR} abs/1807.09956.

\bibitem[\protect\citeauthoryear{Kazemi and
  Elqursh}{2017}]{DBLP:journals/corr/KazemiE17}
Kazemi, V., and Elqursh, A.
\newblock 2017.
\newblock Show, ask, attend, and answer: {A} strong baseline for visual
  question answering.
\newblock {\em CoRR} abs/1704.03162.

\bibitem[\protect\citeauthoryear{Krishna \bgroup et al\mbox.\egroup
  }{2017}]{Krishna2017}
Krishna, R.; Zhu, Y.; Groth, O.; Johnson, J.; Hata, K.; Kravitz, J.; Chen, S.;
  Kalantidis, Y.; Li, L.-J.; Shamma, D.~A.; Bernstein, M.~S.; and Fei-Fei, L.
\newblock 2017.
\newblock Visual genome: Connecting language and vision using crowdsourced
  dense image annotations.
\newblock {\em International Journal of Computer Vision} 123(1):32--73.

\bibitem[\protect\citeauthoryear{Kulesza \bgroup et al\mbox.\egroup
  }{2012}]{kulesza2012tell}
Kulesza, T.; Stumpf, S.; Burnett, M.; and Kwan, I.
\newblock 2012.
\newblock Tell me more?: the effects of mental model soundness on personalizing
  an intelligent agent.
\newblock In {\em Proceedings of the SIGCHI Conference on Human Factors in
  Computing Systems},  1--10.
\newblock ACM.

\bibitem[\protect\citeauthoryear{Lane \bgroup et al\mbox.\egroup
  }{2005}]{lane2005explainable}
Lane, H.~C.; Core, M.~G.; Van~Lent, M.; Solomon, S.; and Gomboc, D.
\newblock 2005.
\newblock Explainable artificial intelligence for training and tutoring.
\newblock Technical report, UNIVERSITY OF SOUTHERN CALIFORNIA MARINA DEL REY CA
  INST FOR CREATIVE~….

\bibitem[\protect\citeauthoryear{Lomas \bgroup et al\mbox.\egroup
  }{2012}]{lomas2012explaining}
Lomas, M.; Chevalier, R.; Cross~II, E.~V.; Garrett, R.~C.; Hoare, J.; and
  Kopack, M.
\newblock 2012.
\newblock Explaining robot actions.
\newblock In {\em Proceedings of the seventh annual ACM/IEEE international
  conference on Human-Robot Interaction},  187--188.
\newblock ACM.

\bibitem[\protect\citeauthoryear{Lu \bgroup et al\mbox.\egroup
  }{2016a}]{lu2016hierarchical}
Lu, J.; Yang, J.; Batra, D.; and Parikh, D.
\newblock 2016a.
\newblock Hierarchical question-image co-attention for visual question
  answering.
\newblock In {\em Advances In Neural Information Processing Systems},
  289--297.

\bibitem[\protect\citeauthoryear{Lu \bgroup et al\mbox.\egroup
  }{2016b}]{DBLP:journals/corr/LuYBP16}
Lu, J.; Yang, J.; Batra, D.; and Parikh, D.
\newblock 2016b.
\newblock Hierarchical question-image co-attention for visual question
  answering.
\newblock {\em CoRR} abs/1606.00061.

\bibitem[\protect\citeauthoryear{Narayanan \bgroup et al\mbox.\egroup
  }{2018}]{narayanan2018humans}
Narayanan, M.; Chen, E.; He, J.; Kim, B.; Gershman, S.; and Doshi-Velez, F.
\newblock 2018.
\newblock How do humans understand explanations from machine learning systems?
  an evaluation of the human-interpretability of explanation.
\newblock {\em arXiv preprint arXiv:1802.00682}.

\bibitem[\protect\citeauthoryear{Pennington, Socher, and
  Manning}{2014}]{pennington2014glove}
Pennington, J.; Socher, R.; and Manning, C.
\newblock 2014.
\newblock Glove: Global vectors for word representation.
\newblock In {\em Proceedings of the 2014 conference on empirical methods in
  natural language processing (EMNLP)},  1532--1543.

\bibitem[\protect\citeauthoryear{Ray \bgroup et al\mbox.\egroup
  }{2019}]{ray2019lucid}
Ray, A.; Burachas, G.; Yao, Y.; and Divakaran, A.
\newblock 2019.
\newblock Lucid explanations help: Using a human-ai image-guessing game to
  evaluate machine explanation helpfulness.
\newblock {\em arXiv preprint arXiv:1904.03285}.

\bibitem[\protect\citeauthoryear{Ribeiro, Singh, and
  Guestrin}{2016}]{ribeiro2016should}
Ribeiro, M.~T.; Singh, S.; and Guestrin, C.
\newblock 2016.
\newblock Why should i trust you?: Explaining the predictions of any
  classifier.
\newblock In {\em Proceedings of the 22nd ACM SIGKDD international conference
  on knowledge discovery and data mining},  1135--1144.
\newblock ACM.

\bibitem[\protect\citeauthoryear{Selvaraju \bgroup et al\mbox.\egroup
  }{2017}]{selvaraju2017grad}
Selvaraju, R.~R.; Cogswell, M.; Das, A.; Vedantam, R.; Parikh, D.; and Batra,
  D.
\newblock 2017.
\newblock Grad-cam: Visual explanations from deep networks via gradient-based
  localization.
\newblock In {\em Proceedings of the IEEE International Conference on Computer
  Vision},  618--626.

\bibitem[\protect\citeauthoryear{Shortliffe and
  Buchanan}{1984}]{shortliffe1984model}
Shortliffe, E.~H., and Buchanan, B.~G.
\newblock 1984.
\newblock A model of inexact reasoning in medicine.
\newblock {\em Rule-based expert systems}  233--262.

\bibitem[\protect\citeauthoryear{Szegedy \bgroup et al\mbox.\egroup
  }{2017}]{szegedy2017inception}
Szegedy, C.; Ioffe, S.; Vanhoucke, V.; and Alemi, A.~A.
\newblock 2017.
\newblock Inception-v4, inception-resnet and the impact of residual connections
  on learning.
\newblock In {\em Thirty-First AAAI Conference on Artificial Intelligence}.

\bibitem[\protect\citeauthoryear{Teney \bgroup et al\mbox.\egroup
  }{2017}]{DBLP:journals/corr/abs-1708-02711}
Teney, D.; Anderson, P.; He, X.; and van~den Hengel, A.
\newblock 2017.
\newblock Tips and tricks for visual question answering: Learnings from the
  2017 challenge.
\newblock {\em CoRR} abs/1708.02711.

\bibitem[\protect\citeauthoryear{Teney \bgroup et al\mbox.\egroup
  }{2018}]{teney2018tips}
Teney, D.; Anderson, P.; He, X.; and van~den Hengel, A.
\newblock 2018.
\newblock Tips and tricks for visual question answering: Learnings from the
  2017 challenge.
\newblock In {\em Proceedings of the IEEE Conference on Computer Vision and
  Pattern Recognition},  4223--4232.

\bibitem[\protect\citeauthoryear{Van~Lent, Fisher, and
  Mancuso}{2004}]{van2004explainable}
Van~Lent, M.; Fisher, W.; and Mancuso, M.
\newblock 2004.
\newblock An explainable artificial intelligence system for small-unit tactical
  behavior.
\newblock In {\em Proceedings of the national conference on artificial
  intelligence},  900--907.
\newblock Menlo Park, CA; Cambridge, MA; London; AAAI Press; MIT Press; 1999.

\bibitem[\protect\citeauthoryear{Xu and
  Saenko}{2015}]{DBLP:journals/corr/XuS15a}
Xu, H., and Saenko, K.
\newblock 2015.
\newblock Ask, attend and answer: Exploring question-guided spatial attention
  for visual question answering.
\newblock {\em CoRR} abs/1511.05234.

\bibitem[\protect\citeauthoryear{Xu and
  Saenko}{2016}]{10.1007/978-3-319-46478-7_28}
Xu, H., and Saenko, K.
\newblock 2016.
\newblock Ask, attend and answer: Exploring question-guided spatial attention
  for visual question answering.
\newblock In Leibe, B.; Matas, J.; Sebe, N.; and Welling, M., eds., {\em
  Computer Vision -- ECCV 2016},  451--466.
\newblock Cham: Springer International Publishing.

\bibitem[\protect\citeauthoryear{Yang \bgroup et al\mbox.\egroup
  }{2016}]{yang2016stacked}
Yang, Z.; He, X.; Gao, J.; Deng, L.; and Smola, A.
\newblock 2016.
\newblock Stacked attention networks for image question answering.
\newblock In {\em Proceedings of the IEEE conference on computer vision and
  pattern recognition},  21--29.

\bibitem[\protect\citeauthoryear{Zeiler and
  Fergus}{2014}]{zeiler2014visualizing}
Zeiler, M.~D., and Fergus, R.
\newblock 2014.
\newblock Visualizing and understanding convolutional networks.
\newblock In {\em European conference on computer vision},  818--833.
\newblock Springer.

\bibitem[\protect\citeauthoryear{Zhou \bgroup et al\mbox.\egroup
  }{2014}]{zhou2014object}
Zhou, B.; Khosla, A.; Lapedriza, A.; Oliva, A.; and Torralba, A.
\newblock 2014.
\newblock Object detectors emerge in deep scene cnns.
\newblock {\em arXiv preprint arXiv:1412.6856}.

\end{thebibliography}
\bibliographystyle{aaai}
\end{document}